\pdfoutput=1

\documentclass[11pt]{article}

\usepackage[]{ACL2023}

\usepackage{times}
\usepackage{latexsym}

\usepackage[T1]{fontenc}

\usepackage[utf8]{inputenc}

\usepackage{microtype}

\usepackage{inconsolata}

\usepackage[utf8]{inputenc}
\usepackage[T1]{fontenc}
\usepackage{mathpazo}
\usepackage{multirow}
\usepackage{tabularx}
\usepackage{comment}
\usepackage{amsmath,amssymb,amsfonts}
\usepackage{algorithmic}
\usepackage{graphicx}
\usepackage{textcomp}
\usepackage{listings}
\usepackage{pdfpages}
\usepackage{tikz}
\usepackage{pgfplots}
\usepackage{hyperref}
\usepackage{cleveref}
\usepackage{color}
\usepackage[linesnumbered,ruled,commentsnumbered]{algorithm2e}
\usepackage{booktabs}
\usepackage{float}
\pgfplotsset{compat=1.18} 

\newcommand\blfootnote[1]{%
  \begingroup
  \renewcommand\thefootnote{}\footnote{#1}%
  \addtocounter{footnote}{-1}%
  \endgroup
}

\title{UnifiedCrawl: Aggregated Common Crawl for Affordable Adaptation of LLMs on Low-Resource Languages}

\author{Bethel Melesse Tessema \\ Ajou University \\  Suwon, South Korea \And           Akhil Kedia \\ Independent Researcher \\ Seoul, South Korea \And
        Tae-Sun Chung \\ Ajou University \\  Suwon, South Korea }

\begin{document}
\maketitle
\begin{abstract}
Large language models (LLMs) under-perform on low-resource languages due to limited training data. We present a method to efficiently collect text data for low-resource languages from the entire Common Crawl corpus. Our approach, UnifiedCrawl, filters and extracts common crawl using minimal compute resources, yielding mono-lingual datasets much larger than previously available sources. We demonstrate that leveraging this data to fine-tuning multilingual LLMs via efficient adapter methods (QLoRA) significantly boosts performance on the low-resource language, while minimizing VRAM usage. Our experiments show large improvements in language modeling perplexity and an increase in few-shot prompting scores. Our work and released source code provide an affordable approach to improve LLMs for low-resource languages using consumer hardware. Our source code is available here at \url{https://github.com/bethelmelesse/unifiedcrawl}.

\end{abstract}
\blfootnote{Correspondence: bethelmelesse01@gmail.com}
\section{Introduction}
\label{sec:intro} 

Generative AI has become an integral part of our daily lives, assisting us in various ways, whether it is through its Natural Language Processing (NLP) or Computer Vision (CV) specialty. 

In the field of Natural Language Processing (NLP), generative models play a pivotal role in generating coherent and contextually relevant text. They leverage deep learning architectures, particularly transformer-based architectures, and are pre-trained on vast amounts of textual data to learn the nuance of language. These models learn the patterns and structures of language from large datasets, allowing them to generate new text similar to the input data. 

These models, fueled by Large Language Models (LLMs), are characterized by their extensive size, often measured in terms of the number of parameters, often many billions. This immense number of parameters allows these models to capture complex language patterns and context, resulting in improved performance on various NLP tasks.

For example, OpenAI's GPT (Generative Pre-trained Transformer) series \cite{gpt3,chatgpt,gpt35turbo,gpt4} has played a fundamental role in transforming the public’s view and usage of AI NLP tools. GPT-3~\cite{gpt3}, with its staggering 175 billion parameters, represents a groundbreaking milestone, showcasing the scalability of transformer-based models. This technology has shown a substantial economic potential due to its broad applicability in commercial usage.

\subsection{Problem Definition}

By leveraging the availability of extensive large and diverse dataset, often composed of high-resource languages, LLMs have shown remarkable performance in generating content within those linguistic contexts, mimicking human-like responses. However, their performance diminishes significantly when prompted with low-resource languages due to the limited availability of training data and resources for those languages. This limitation results in the generation of responses that lack coherence. For example, when prompted with queries in a low-resource language (such as Amharic (ISO:amh), the most widely language in Ethiopia) models like GPT-turbo-3.5\cite{gpt35turbo} produce incomprehensible outputs. This challenge persists even when inputting prompts in high-resource languages and instructing the model to respond in low-resource languages, resulting in sentences that lack meaningful coherence. 

The limitation of LLMs in handling low-resource languages stems from their initial training which heavily relies on vast amounts of primarily English-centered data. \Cref{appendix:lang_dist1,appendix:lang_dist2} illustrates the distribution of data and the percentage constituting high-resource and low-resource languages in the training process for these LLMs. 

Addressing this challenge of adapting LLMs for use in low-resource languages is crucial for democratizing their accessibility and broadening their practical applicability.  However, pre-training LLMs can be exceptionally costly, primarily for two main reasons.

First, as mentioned earlier, pre-training LLMs requires an extensive amount of textual data, and low-resource languages often lack the resources to meet this requirement. For instance, in widely used collection Common Crawl (CC)~\cite{commoncrawl}, low-resource languages such as Tagalog, Punjabi, Kurdish, Lao, Amharic, etc., constitute a minuscule fraction (less than 0.01\%) compared to other high-resource languages like English, German, and Russian (\ref{appendix:A}).

Second, the resource-intensive nature of training LLMs, characterized by an extensive number of parameters, demands substantial GPU power, memory, and time. For example, models like gpt-3.5-turbo (175 billion parameters), Claude~\cite{claude} (52 billion parameters), and LLaMA~\cite{llama} (1.6 billion parameters) translate into an exceedingly resource-intensive training process. \Cref{tab:llm_size} provides the size details of these LLMs. Consequently, the immense size of these LLMs renders training prohibitively expensive and inaccessible to non-wealthy communities/nations, smaller companies, and educational institutions.

In this paper, our primary objectives are to investigate the following research questions:

\begin{enumerate}
    \item How can we enhance LLMs to  perform well in low-resource languages?
    \item How can we collect sufficient training data in low-resource languages for LLMs?
    \item How can we achieve the above, while being constrained by consumer devices' memory, storage, and compute?
\end{enumerate}

\begin{figure}[H]
	\centering
	\includegraphics[width=\columnwidth]{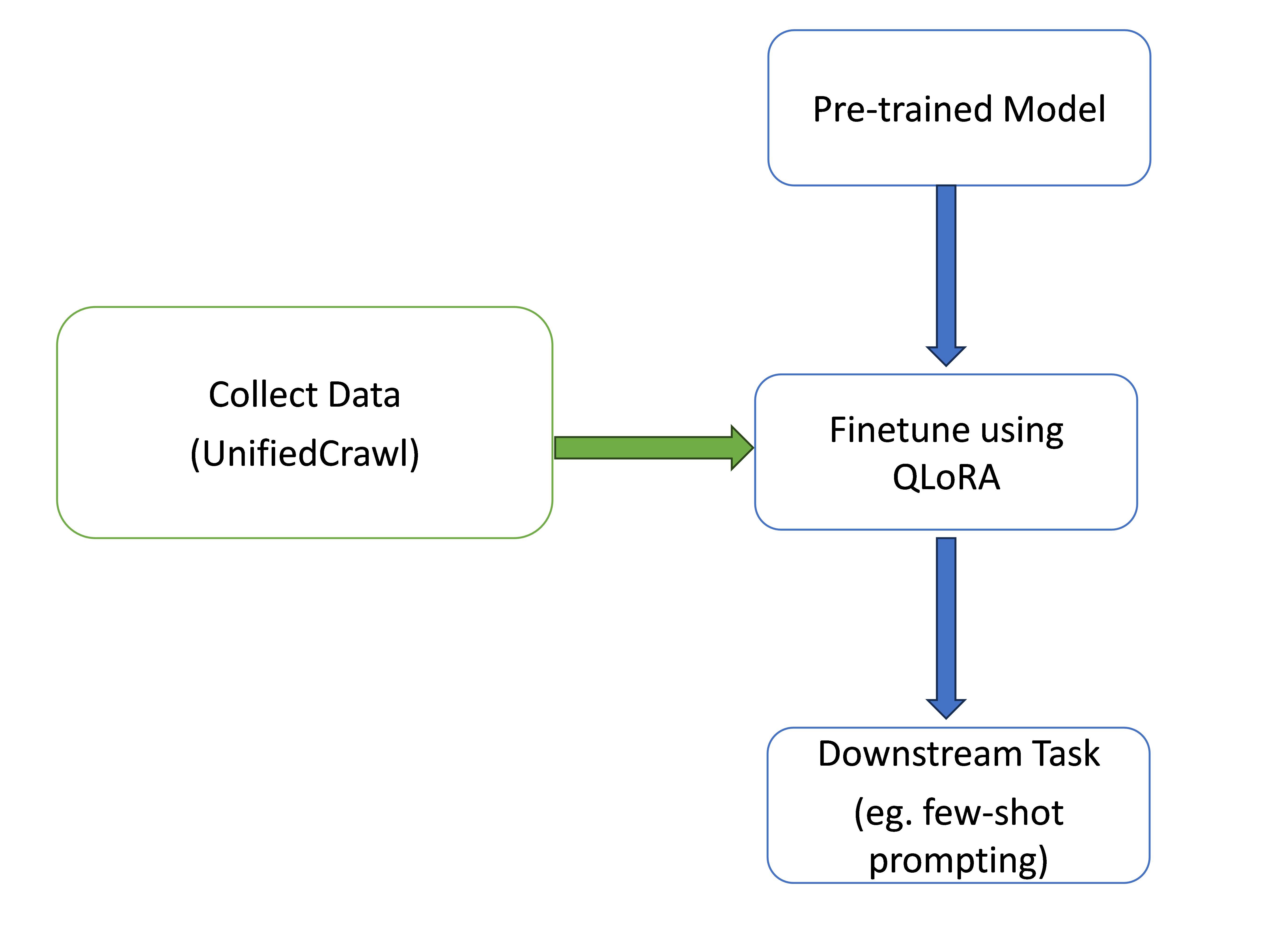} 	
	\caption{\small \centering {High-Level Overview of our Method}}
	\label{fig:highlevel}
\end{figure}

\subsection{Proposed Method}

To address the aforementioned challenges, we present a novel approach aimed at overcoming data scarcity for low-resource languages, and leverage efficient methods for training LLMs on low-cost hardware. 

\begin{figure*}[!htb]
	\centering
	\includegraphics[width=1.0\textwidth]{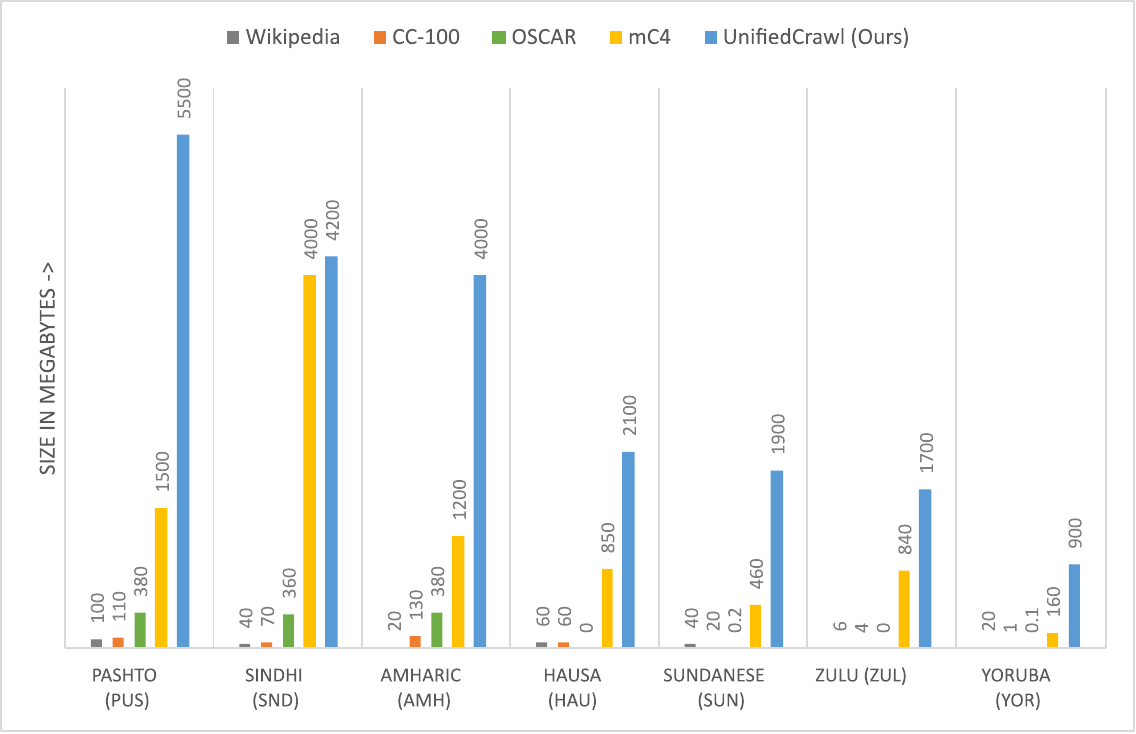} 	
	\caption{\small \centering {Our Dataset is much Larger than all Prior Works}}
	\label{fig:dataset_comparision}
\end{figure*}

Our proposed method involves the development of  an efficient and cost-effective data collection strategy to extract comprehensive textual content specific to a given low-resource language from the entire Common Crawl corpus. \Cref{fig:data_collection} illustrates our architecture. By carefully paying particular attention to memory, compute and network usage in each step of our data collection pipeline, our method is optimized to run entirely on personal consumer hardware - the entirety of the Common Crawl dataset can be achieved in a matter of days, utilizing less than 10GB of RAM and storage. The outcome of this process is our meticulously curated dataset called UnifiedCrawl. Using our method, we were able to successfully extract a monolingual corpora for specific low-resource languages, significantly surpassing the sizes of previously compiled collections, as shown in \cref{fig:dataset_comparision}. 

Subsequently, we leverage quantization and lightweight low-rank adapters for fine-tuning multilingual Large Language Models (LLMs) on the collected dataset. This innovative technique facilitates the utilization of exceptionally large models on consumer-grade GPUs, thereby enhancing accessibility and affordability in training. 

\Cref{fig:highlevel} illustrates the overarching concept of our proposed scheme. Our approach involves fine-tuning a pre-trained model on our UnifiedCrawl dataset, extracted from the common crawl corpus using our data extraction method. The resulting fine-tuned model can then be applied to downstream tasks.

\section{Related Works}
\label{chapter 2}

\label{sec:relatedworks}

\subsection{Multilingual Large Language Models}
In recent years, there has been a notable surge in the development of multilingual Large Language Models (LLMs), contributing to improved cross-lingual understanding. Examples of these large language models are shown in the \cref{appendix:llm_size}, including their model type, size and the number of languages they are trained on.

While these multilingual models have made strides towards linguistic inclusivity by covering over many languages, they still overlook hundreds of lower-resource languages with sizable speaker populations. This hinders the efficacy of models in many languages compared to performance in high-resource languages. This limitation is primarily due to the lack of sufficient online training data available for lower-resource languages.

In this work, we aim to improve the performance of the above-mentioned models (particularly XGLM model) in low-resource languages by training them on our collected dataset.

\subsection{Large Multilingual or Monolingual Datasets}

We have noted that data is a crucial component for training language models specifically in the multilingual domain. However, there is a considerable gap in data quantity across languages and domains. Even within the largest Common Crawl corpus, which is a vast web archive that encompasses a diverse collection of web pages providing a rich source of textual data in a multitude of languages and topics, over 41 languages make up less than 0.01\% of the data, and 100 languages 0.1\% - The quantity of data in common crawl decreases almost exponentially as shown in \cref{fig:lang_dist1,fig:lang_dist2}. This leaves only a handful of the world’s languages represented in evolving language technologies and applications~\cite{inclusivity}. 

In this study, we extract all the available textual data from every archive within the common crawl for a specific low resource language. Our choice of the common crawl dataset is driven by our objective to maximize the acquisition of the available data, leveraging its vast size and inclusivity of many languages, as it systematically scraps data across the internet.

\subsection{Common Crawl and Dataset Extraction}

Due to the extensive scope, Common Crawl (CC) and its derivatives are often used to pre-train large language models, and the majority of State-of-the-art models, such as LLaMA~\cite{llama} , GPT-3~\cite{gpt3} , Falcon~\cite{falcon} , PALM~\cite{palm} , Stable LM~\cite{stablelm} , etc. have incorporated datasets sourced from the Common Crawl corpus into their training pipelines. This integration has contributed to their improved proficiency in understanding and generating human-like text across various domains and languages. 

Several smaller datasets have been extracted from the Common Crawl corpus, each contributing to the training of language models. Examples include CC-net~\cite{ccnet}, which extracted monolingual corpora for transformer model training; mC4~\cite{mc4}, which collected data from publicly accessible CC archives; and OSCAR project~\cite{oscar}, which focuses on releasing monolingual corpora from recent CC archives. These subsets have then been used to train many of the State-of-the-art models, such as mT5 (used mC4)~\cite{mT5}, BLOOM (used OSCAR)~\cite{bloom}, etc. 

However, a common issue persists: many extracted datasets from the common corpus are often limited to one language (eg. CC-net), or a few archives(eg. OSCAR), or are not updated with latest common crawl dumps (eg. mC4). Moreover, due to the sheer scale of the corpus, naively extracting text data for a specific language from all the common crawl archives can be challenging, as it can be time and memory intensive.  Moreover, these datasets can also not be easily updated with more data from latest common crawl archives. This limitation hinders the extraction of data for specific languages, especially for very low-resource languages, contributing to the lack of linguistic diversity in the available datasets. 

In response to these challenges and limitations, we present a cost-effective means of extracting text data from all CC archives for low-resource languages, including the latest common crawl archives which are much larger compared to previous archives. Our contribution includes releasing the code base for other fellow researchers to extract their own dataset from CC for low resource languages. By doing so, we aim to address the existing gaps in dataset extraction methodologies and contribute to the advancement of linguistic research in low-resource language contexts.

\subsection{Deduplication} 

Another method adopted in our work includes deduplication techniques. Raw text datasets obtained through web scraping often contain the same lines multiple times~\cite{deduplicating}. This repetition within the dataset can negatively affect the learning process as it slows down the training as well as limits the model's generalization capabilities. To overcome these challenges, it is important to apply some form of deduplication on the extracted dataset.

Numerous deduplication methods have been previously proposed and employed in prior works. CC-Net, for instances, utilized a paragraph-based Exact-Hash deduplication, whereas other approximate methods, such as MinHash~\cite{minhash}, MinHash-LSH~\cite{minhashlsh}, SimHash~\cite{simhash1,simhash2}, are sometimes used for faster deduplication in different context~\cite{bloom, falcon}. 

In our data extraction pipeline, we opted for~\cite{deduplicating}'s exact substring deduplication method – same approach adopted in mC4, OSCAR, and CC100. This approach not only effectively addresses redundancy but also removes common header/footer artifacts often present in the extracted text, enhancing the overall quality of the dataset. By employing this deduplication method within our proposed scheme, our goal is to extract a high-quality dataset that contributes positively to model’s training, resulting in accelerated training, improved perplexity, and reduced likelihood of model memorization. 

\subsection{Low Resource Model Adaptation}

Training (pretraining/fine-tuning) large language models is often impractical beyond major corporate or institutional settings due to their substantial parameter count and the resource-intensive nature of training LLMs. For instance, training a model with a large number of parameters consumes considerable GPU memory and time - for example a 7B model requires 28GB of GPU VRAM, outside the scope of most consumer GPUs.
An effective solution to mitigate this challenge involves the integration of quantization techniques into LLMs. Quantization can be achieved through methods such as Quantization-aware training~\cite{bitnet} or post-training quantization approaches like GptQ ~\cite{gptq}, SmoothQuant ~\cite{smoothquant}, bitsandbytes ~\cite{bitsnbytes}, among others. These techniques work by reducing the precision of model parameters, allowing for more efficient storage and computation, dramatically reducing GPU VRAM usage.

However, fine-tuning still demands expensive gradients over model parameters. To address this, a more resource-efficient approach involves training adapters on frozen LLMs, known as Low-Rank adaptation (LoRA), proposed by~\cite{lora}. LoRA strategically freezes pre-trained model weights and introduces a smaller trainable weight into the model’s architecture. As only these added low-rank matrices are trained, there is significant reduction of the overall trainable parameter count and optimizer states, and correspondingly of GPU memory requirement. Experiments on several pretrained models, such as RoBERTa~\cite{roberta}, DeBERTa~\cite{deberta}, GPT-2~\cite{gpt2}, GPT-3~\cite{gpt3}, have shown that LoRA achieves comparable or even better performances than existing adapter and prompt-based methods.

This method was further extended to QLoRA~\cite{qlora}, which combines quantization with adapter training. QLoRA achieves a further reduction in memory usage, enabling the fine-tuning of a 65-billion-parameter model on a single 48GB GPU while maintaining full 16-bit fine-tuning task performance. 

As QLoRA consistently delivers performance similar to full fine-tuning ~\cite{qlora_2, qlora_3, qlora_4, qlora_5} for much lower VRAM, we adopt QLoRA in our work to balance computational efficiency with model performance.

\section{Methods}
\label{Chapter 3}
\label{sec:methods}

In this section, we first present a method and procedure to collect and process training data for low-resource languages from the common crawl dataset using limited computing resources. Additionally, we adopt a method to efficiently train large language models on the extracted training dataset using limited GPU resources.

\begin{figure}[H]
	\centering
	\includegraphics[width=\columnwidth]{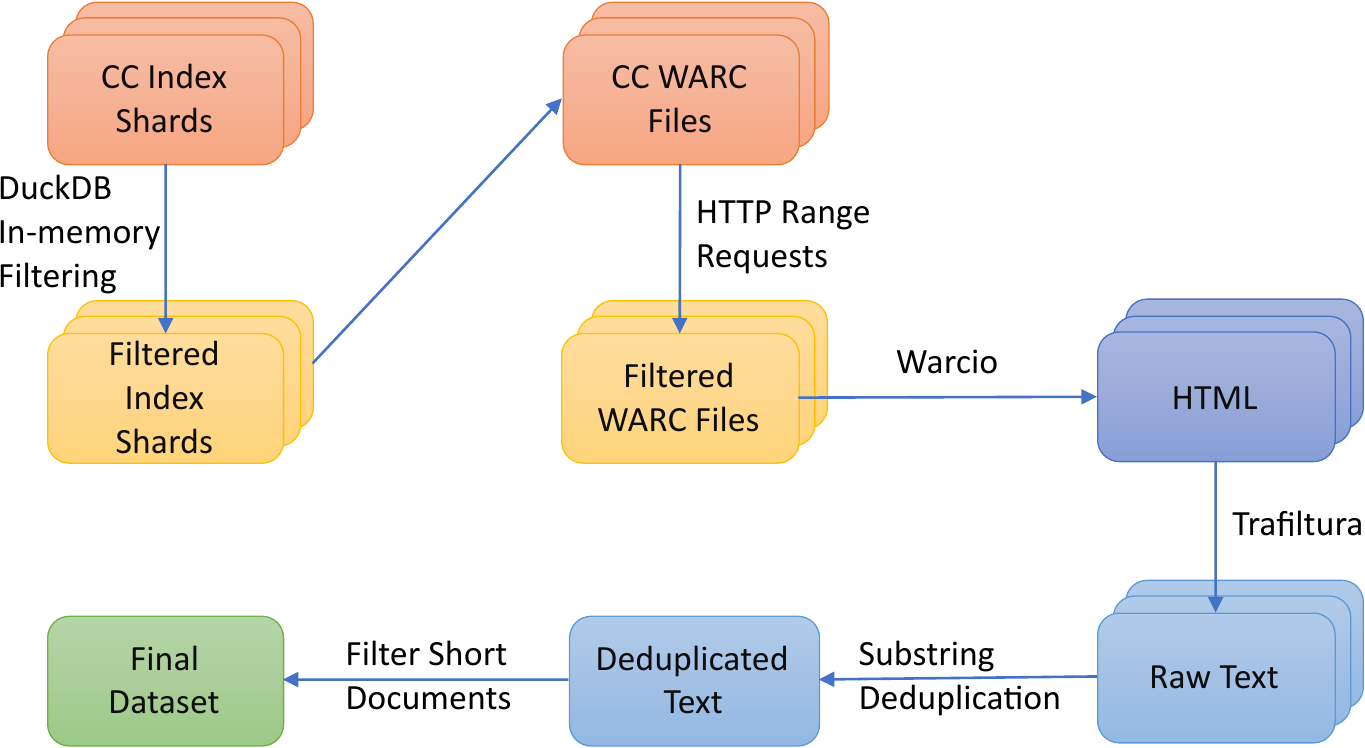} 	
	\caption{\small \centering {UnifiedCrawl: Data Extraction Framework}}
	\label{fig:data_collection}
\end{figure}

\subsection{Data Collection Framework}

The raw text data for low-resource languages is gathered from the common crawl dataset. The common crawl dataset is extremely large, at approximately 100 TeraBytes per crawl archive, with multiple archives per year\footnote{\url{https://commoncrawl.org/blog/jan-feb-2023-crawl-archive-now-available}}. Due to its sheer size, it is difficult to directly download raw text dataset from the corpus. In this subsection, we propose an efficient and cost-effective framework to extract data from a single common crawl archive, which is repeated for all available dumps. \Cref{fig:data_collection} illustrates our data collection pipeline for extracting raw text dataset from a single common crawl archive.

\subsubsection{Index Filtering}
The Common Crawl provides a columnar index (CC index\footnote{\url{https://commoncrawl.org/blog/index-to-warc-files-and-urls-in-columnar-format}}) containing language annotations\footnote{\url{https://commoncrawl.org/blog/august-2018-crawl-archive-now-available}} for each URL in the archive. We exploit this information to selectively extract a specific low-resource language from a single archive (eg. CC-MAIN-2023-23). However, even this CC index is typically 100s of GBs for a single archive. As we process 43 archives, this would result in a total of 10s of TBs of just the index, which would be difficult to store.

Instead, we utilize DuckDB~\cite{duckdb}, an in-memory analytical database management system, to filter the index shards corresponding to our target language as the primary content language. By employing DuckDB's in-memory filtering and downloading, we eliminate the need for storage-intensive initial bulk downloads and subsequent filtering.

Additionally, we integrate python’s multiprocessing package with DuckDB to further reduce the overall downloading time. This integration utilizes multiple processes concurrently across all CPU cores on a single system, leveraging parallelism and bypassing the global interpreter lock to expedite the data acquisition process. The combined utilization of DuckDB and multiprocessing significantly optimizes storage usage and accelerates the overall downloading process.

\subsubsection{Extracting WARC Files}
The filtered and downloaded index shards from the previous step contains the path to the WARC (Web ARChive) files~\cite{warc}, which contains the content of the crawled webpages and its metadata. Due to the considerable size of the CC archive shards containing these WARC files, downloading all WARC files is impractical. Instead, we selectively filter and retain WARC files corresponding to our target languages within the index shards – avoiding the download of all the WARC files. 

This filtering and downloading process utilizes the columnar information provided in the index shard that include the WARC filename, WARC record offset, and WARC record length (the URL and Content Languages we leveraged earlier during the index filtering step are also present here). The WARC filename gives a URL to the CC archive shard containing this particular WARC file, the WARC record offset indicates the exact location of the WARC file we need, the WARC length specifies its span. 

Leveraging this information, we download the corresponding WARC for each URL containing our target language as its primary content language via HTTP Range Requests~\cite{range}. This method of downloading allows us to download only the WARC files that we need and skip the rest from downloading. This is done by requesting the server to send only a portion of an HTTP message back to the client -- in our case, only the files corresponding to our target language. By only downloading these necessary WARC files, we conserve bandwidth and storage that result from downloading all the WARC files.

\subsubsection{Text Extraction}
The next step involves extracting the raw text. We start by retrieving the HTML source from the downloaded WARC files by using the WARCIO library~\cite{warcio}, which offers a fast and efficient way to read and write WARC format. From this extracted html source, we finally obtain the raw text using the command line tool Trafilatura ~\cite{trafilatura} (same tool as used in~\cite{refinedweb}), which is specifically designed to extract text from the web. This library not only facilitates the extraction of the text from HTML but also improves the text quality by eliminating the noise caused by recurring elements such as headers, footers and other information. 

It is noteworthy that the entire process, including the downloading of WARC files as well as the reading and the text extraction, is conducted in-memory for the purpose of reducing the time and storage requirement. By avoiding storing unnecessary elements found within the warc files and the raw html inside the warc (such as large quantities of javascript/html tags/etc), we dramatically reduce the storage requirements.

\subsubsection{Deduplication}
While the Trafilatura library improves the quality of our extracted text, it is common to have some repetitive sequences within the raw text data. These repetitive sequences include things like copyright notices, some headers/footers, keywords, etc. that are common in many similar websites across its pages. Having multiple repetitive sequence reduces the overall quality of the training data as it encourages the model to prioritize memorization rather than make the model learn to generalize. Therefore, to further improve the quality of our data as well as improve the training process (later on), we remove these deduplicated elements from the documents. We adopted exact substring deduplication~\cite{deduplicating} technique to remove all duplicate occurrences over a given length from the dataset. However, after removing the repeated sequences, some documents become too short -- we hence discard documents of very short length. This final step yields to our final dataset, which we call UnifiedCrawl.

\subsection{Low Resource Model Adaptation}
Common multi-lingual Large Language Models (LLMs) often suffer from low performance on the long-tail of low-resource languages(~\cite{xglm}, and \cref{results:ppl}). Finetuning LLMs on our dataset offers a pathway to democratize AI, improving LLMs on low-resource languages. We hence focus on the regime of using consumer hardware. 

LLMs require large GPU memory to simply store the parameters, limiting the maximum model size we can train/infer. Using 4-bit quantization~\cite{bitsnbytes}, we can fit almost 3-4x larger models in the same GPU memory, compared to using 16-bit parameters, at the cost of some precision. The performance improvements of a 4x larger LLM compared to a smaller one 16-bit precision is much larger than the slight loss caused by reduced precision, hence it is beneficial to use as large a model as we show in \cref{results:method}.

Furthermore, finetuning LLMs would also require large GPU memory to store gradients and optimizer states for all parameters. We instead leverage Quantized Low-Rank Adapters (QLoRA, \cite{qlora}) to efficiently train adapters on the quantized LLMs. This significantly reduces memory usage without compromising performance, enabling training of much larger models. Using QLoRA with larger models outperforms full finetuning of smaller models as we will show in \cref{sec:ablation:full_finetuning}.

Our data extraction method results in datasets larger than prior datasets sourced from the common crawl. We also show that our model adaptation method results in large improvements on Language Modeling perplexity, and on few-shot prompting~\cite{gpt3} on downstream Question-Answering tasks (\cref{res:downstream}) and outperforms full-finetuning of smaller models.

\section{Experimental Settings and Implementation Details}
	\label{Chapter 4}

\label{sec:setup}

\subsection{Languages and Benchmark Datasets and Dataset Collection}

\subsubsection{Dataset Collection}

Data collection of UnifiedCrawl was carried out using consumer-grade $500$MBps internet connection. Our extracted raw text was formatted in the HuggingFace~\cite{huggingface} dataset format. Since the substring deduplication method of \cite{deduplicating} cannot directly handle datasets in this format, we utilized Text-dedup~\cite{text_dedup} to wrap~\cite{deduplicating}’s implementation of deduplication for compatibility with HuggingFace format. We removed all duplicate substrings of length at-least $50$ and all documents shorter than $100$ characters, a decision made arbitrarily but following the approach of~\cite{refinedweb}. 

\subsubsection{Compute Requirements} 
Index filtering is constrained by network download bandwidth for any low-resource language, as the majority of the index is discarded, filtering down to $10$s of MBs. The index for all the archives can be processed in a few days on a consumer internet connection of $500$MBps using $<10$GB of RAM. Alternatively, using a single AWS server with 12Gbps network, each archive can be processed in $<20$ minutes, and the entire CC filtered for $<4$USD in $<1$ day. Cloud big-data querying services such as AWS Athena\footnote{\url{https://aws.amazon.com/athena/}} can run this step much faster, but at the cost of $100$s of USD. Text extraction and de-duplication from an archive can be processed in only a few minutes, and all CC archives can be processed in a few hours.

\subsubsection{Languages} 
Our data extraction method underwent testing on seven languages: Hausa (hau), Pashto (pus), Amharic (amh), Yoruba (yor), Sudanese (sun), Sindhi (snd), and Zulu (zul), ordered by the descending number of speakers. We specifically selected very low-resource languages (that constituted less than 0.004\% of the Common Crawl dataset each), with the highest number of speakers. \Cref{tab:num_speakers} provides details on each language's ISO code, corresponding number of speakers (in millions), their representation percentage in the Common Crawl dataset (for the ` CC-MAIN-2023-14'' archive), and the geographical region where these languages are spoken. By applying our method to these languages, we aim to demonstrate that our implementation and approach are language-agnostic.

\begin{table*}[htbp]
\centering
\begin{tabular}{ l c c l }
\toprule
\textbf{Language (ISO)} & \textbf{Fraction of CC} & \textbf{\# Speakers(M)} & \textbf{Geographical Region} \\
\midrule
Hausa (hau) & 0.0036\% & 80 & Nigeria, Chad, Cameroon, Ghana \\
Pashto (pus) & 0.0033\% & 60 & Afghanistan, Pakistan \\
Amharic (amh) & 0.0036\% & 60 & Ethiopia \\
Yoruba (yor) & 0.0011\% & 50 & Benin, Nigeria, Togo \\
Sundanese (sun) & 0.0011\% & 40 & Indonesia \\
Sindhi (snd) & 0.0017\% & 30 & Pakistan, India \\
Zulu (zul) & 0.0016\% & 30 & South Africa, Lesotho \\
\bottomrule
\end{tabular}
\caption{Details of 7 languages used for Data Collection Evaluation}
\label{tab:num_speakers}
\end{table*}

\subsubsection{Benchmark Datasets} 
To assess the scale and efficacy of our UnifiedCrawl dataset, we conducted comparative analysis of its size against other notable datasets sourced from common crawl. The dataset included in this benchmarking are OSCAR, mC4, CC-100 and Wikipedia. This comparative evaluation aims to provide insights into the relative size and representativeness of Unifiedcrawl in comparison to these widely used datasets.

\subsection{Models and Model Adaptation Settings}

\subsubsection{Models}
Given the language expertise of the authors of this thesis, particularly in Amharic, we focused our model adaptation and evaluation on datasets in this specific language. Following the extraction of UnifiedCrawl-Amharic from the Common Crawl Corpus using our method, we fine-tuned a multilingual large language model using the lightweight adapter QLoRA. Among the available pre-trained multilingual large language models, we chose the XGLM model~\cite{xglm} for adaptation. This XGLM Model is available in two size variants: $564$M and $4.5$B models. 

This choice of model is due to its inclusion of the language Amharic in its pretraining dataset. However, this requirement only applies to the XGLM-$4.5$B parameter model. The XGLM-$564$M does not include Amharic in its training data. However, we still explored the adaptation process on the smaller model as well. This deliberate selection enables us to explore and analyze the nuances of the adaptation process, considering the variations in language inclusion within the same model. Furthermore, XGLM is larger in size than mGPT, and it performs equally or better than BLOOM~\cite{bloomplusone}.

\subsubsection{Model Adaptation} 
We use the HuggingFace~\cite{huggingface} library to implement our code base. We use $r=2$ for LoRA rank, as\cite{lora} found small values of r to be effective, and train adapters on all Linear matrices. We finetune these models on our UnifiedCrawl-Amharic dataset for $1$ epoch. While multiple epochs should yield better performance, we only train for $1$ epoch due to compute constraints. We used original/standard hyper-params wherever applicable and did grid search for learning rate. All experiments were carried out on an Nvidia RTX3070 or a RTX3090, and finetuning took $1$ day.

\subsection{Evaluation Settings} 
Our model was evaluated under two settings -- in terms of language modeling, and downstream few-shot prompting. 

\subsubsection{Language Modeling Evaluation} 
For evaluating the model’s capabilities, we compare the perplexity of our model during fine-tuning using QLoRA~\cite{qlora} on our UnifiedCrawl-Amharic dataset and the original XGLM model, for both variants. Perplexity is defined as the exponential of negative cross-entropy of language modeling, and is a measure of how well our language model is doing in predicting the next word in a sequence given the previous ones. Lower perplexity implies the model is becoming better at predicting the next word in a sequence. Perplexity provides a quantitative and direct measure for comparing different models. 

\subsubsection{Downstream Evaluation} 
Testing the language model on downstream tasks is necessary to evaluate the model’s practical applicability, generalization capabilities, and task-specific performance in real-world scenarios. We test our finetuned model on our UnifiedCrawl-Amharic dataset on a downstream task – in order to evaluate the effectiveness of the fine-tuning process. It helps us to know whether our model has learned useful representations during the fine-tuning process, and that it can be applied to diverse tasks beyond language modeling task, including those it wasn’t explicitly trained on.

\subsubsection{Question Answering Task} 
We chose question Answering, which is a task of generating a response given a question and a context, for evaluating our method's performance on downstream application. QA tasks are valuable downstream evaluations for pre-trained language models as they assess the model's comprehension, reasoning abilities, and contextual understanding. Therefore, by evaluating on QA tasks, we can evaluate how well a language model can extract and synthesize information from the context provided, infer relationships between different parts of the text, and generate coherent responses for a given query. We use the AmQA dataset~\cite{amqa} for evaluating the model performance on a downstream Question-Answering task. 

\subsubsection{Few Shot Prompting Evaluation} 
This downstream Question-Answering was done under the few-shot prompting~\cite{gpt3} setting, where the model is given only a small set of examples and is expected to generate the output. This is to assess whether our model can generalize and adapt quickly to new or unseen scenarios, given limited information. For few-shot evaluation on AmQA test set, we use $10$ random Context-Question-Answer examples in the prompt. This number was chosen because more examples in the prompt will simply get truncated due to sequence length limitations. We chose these examples from the AmQA train set ,and we appended a question and context to this prompt chosen from the test sample. Our aim is to generate an answer for the question that is selected from the test set. The closer the answer generated to the ground truth label the better. This few-shot evaluation roughly takes $30$ minutes for $4.5$B XGLM model. 

\subsubsection{Evaluation Metrics} 
We used F1 and EM(Exact Match) scores to evaluate the overall quality and accuracy of our model, as commonly used in Question Answering tasks. F1, which is the harmonic mean of precision and recall, provides a more nuanced evaluation, considering the partial overlaps between the generated answers and the ground truth. Complementing F1 score, the EM score gives the percentage of prediction that exactly matches the ground truth answer. We provided a detailed performance assessment in the next chapter.

\section{Performance Evaluation}
\label{Chapter 5}
\label{sec:eval}

We present analysis of the UnifiedCrawl-language dataset extracted using our data extraction pipeline. We then show experimental results and analysis of the XGLM models fine-tuned on UnifiedCrawl-Amharic using QLoRA. We evaluate the adapted models based on language modeling perplexity and downstream few-shot prompting performance on Question-Answering on AmQA.

\subsection{Data Collection Evaluation}

We processed a total of $43$ archives, starting from “CC-MAIN-2018-43”, which marks the first archive to have language annotations\footnote{\url{https://commoncrawl.github.io/cc-crawl-statistics/plots/languages}}. Using our proposed data collection approach, we collected a monolingual dataset for $7$ languages. This includes Hausa (hau), Pashto (pus), Amharic (amh), Yoruba (yor), Sundanese (sun), Sindhi (snd) and Zulu (zul), chosen based on the number of speakers vs latest crawl size. 

In the following subsection, we provide a detailed analysis focused on the Amharic (amh) language. The final dataset sizes extracted from the Common Crawl for all seven languages are presented in \cref{tab:dataset_size}.

\subsubsection{UnifiedCrawl Amharic}
\textbf{Index Filtering:} Amharic (ISO: amh) is approximately $0.0036\%$ of Common crawl\footnote{\url{https://commoncrawl.github.io/cc-crawl-statistics/plots/languages}}. Each Common Crawl archive index is $\approx 250\mathrm{GB}$ compressed. Hence, the expected size of filtered index should be $0.0036\%*250\mathrm{GB} \approx 10 \mathrm{MB}$ (single archive percentage in the CC * size of archive index). The index filter process resulted in $\approx 20\mathrm{MB}$ of filtered index uncompressed, as expected. We only keep URLs with the only language as our target language to increase the dataset quality as well as speed up the process. Keeping URLs with any occurrence increases the size of the filtered index by $3x$.

\textbf{Extracting WARC files:} Each archive has $\approx 100\mathrm{TB}$ of compressed WARC files. We only download WARCs corresponding to the target language using Range requests, downloading $\approx 3.5\mathrm{GB}$ WARC per archive.

\textbf{Final Text Extraction:} Extracting plaintext from the WARC HTML reduces the size down to $90\mathrm{MB}$, yielding our final total dataset size of $4\mathrm{GB}$ for all archives. 

\textbf{Deduplication:} Sub-string deduplication is first performed within each archive, and then across all archives. Within each archive, deduplication reduces the size by $60\%$ to $40\mathrm{MB}$, $1.8\mathrm{GB}$ across all archives. This is de-duplicated to provide our final dataset of size $600\mathrm{MB}$. Combined, the two de-duplication reduced the dataset size by $85\%$.

\subsubsection{UnifiedCrawl for other Languages}
Similarly, we provide the final sizes of our UnifiedCrawl datasets across 7 languages in \cref{tab:dataset_size}. The first column indicates the languages for which we extracted the datasets, the second column provides the size of the datasets extracted with the primary language exclusively in the content (e.g., content\_language=[amh]), and the third column estimates the size of datasets where the primary language is our target language but also includes some minor content from other languages (e.g., content\_language=[amh, en,..]). Allowing pages with minor content in other languages should increase the dataset size significantly, and we verified this for Yoruba (yor). The size for other languages are estimated based on the fraction of URLs containing other minor languages.

\begin{table}[H]
\centering

\begin{tabular}{ l r r }
\toprule
\textbf{Languages (ISO)} & \textbf{Size} & \textbf{Max Size} \\
\midrule
Hausa (hau) & 2.1 & 7 \\
Pashto (pus) & 5.5 & 20 \\
Amharic (amh) & 4.0 & 24 \\
Yoruba (yor) & 0.9 & 2 \\
Sundanese (sun) & 1.9 & 6 \\
Sindhi (snd) & 4.2 & 15 \\
Zulu (zul) & 1.7 & 6 \\
\bottomrule

\end{tabular}
\caption{UnifiedCrawl-Language Dataset Size. The Size and Max Size are in GBs}
\label{tab:dataset_size}

\end{table}

\subsubsection{Dataset Comparison with other Datasets}

Using our method, we were able to extract monolingual corpora that exceeds the size of other prior art for low-resource languages, often by multiple orders of magnitude. 

For example, our extracted dataset (UnifiedCrawl-Amharic) surpasses the sizes of previous datasets for the Amharic language. To illustrate, Amharic Wikipedia dataset is $22\mathrm{MB}$\footnote{Amharic Wikipedia at TF datasets: \url{https://www.tensorflow.org/datasets/catalog/wikipedia\#wikipedia20230601am}}, the Amharic News Corpus~\cite{amharicnews} is $150\mathrm{MB}$, OSCAR~\cite{oscar} is $500\mathrm{MB}$, and mC4~\cite{mc4} is $1.2\mathrm{GB}$. In contrast, our dataset amounts to 4GB before the deduplication step.

Similarly, we show comparison of the size of our UnifiedCrawl-Language dataset to other prominent datasets, OSCAR\footnote{OSCAR Dataset Size: \url{https://huggingface.co/datasets/oscar}}, mC4\footnote{mC4 Dataset Size: \url{https://github.com/allenai/allennlp/discussions/5265}} , CC-100\footnote{CC-100 Dataset Size: \url{https://data.statmt.org/cc-100/}}, and Wikipedia\footnote{Wikipedia Dataset Size: \url{https://www.tensorflow.org/datasets/catalog/wikipedia}}  in \cref{tab:dataset_comparision}. All sizes in this table are in MB. OSCAR, mC4, CC-100 are datasets sourced from the Common Crawl Corpus, whereas Wikipedia dataset is a collection of cleaned articles of all languages built from the Wikipedia dumps\footnote{\url{https://dumps.wikimedia.org/}} using Tensorflow Datasets.

\begin{table*}[htb]
\centering
\begin{tabular}{ l c r r r r }
\toprule
\textbf{Languages (ISO)} & \textbf{OSCAR} & \textbf{mC4} & \textbf{CC-100} & \textbf{Wikipedia} & \textbf{UnifiedCrawl} \\
\midrule
Hausa (hau) & - & 850& 60& 60& \textbf{2100}\\
Pashto (pus) & 380& 1500& 110& 100& \textbf{5500}\\
Amharic (amh) & 380& 1200& 130& 20& \textbf{4000}\\
Yoruba (yor) & 0.1& 160& 1& 20& \textbf{900}\\
Sundanese (sun) & 0.2& 460& 20& 40& \textbf{1900}\\
Sindhi (snd) & 360& 4000& 70& 40& \textbf{4200}\\
Zulu (zul) & - & 840& 4& 6& \textbf{1700}\\
\bottomrule

\end{tabular}
\caption{Size of UnifiedCrawl-Language vs. Prior Works}
\label{tab:dataset_comparision}

\end{table*}

\subsection{Method Evaluation}
\label{results:method}
We evaluate the performance of the models fine-tuned using QLoRA models on our UnifiedCrawl dataset in two settings -- first, we compare their language modeling capability measured through perplexity (PPL) in upstream, and second, we evaluated the model on downstream few-shot prompting tasks. For both cases, we take the original model as a baseline.

\subsubsection{Language Modeling Evaluation}
For evaluating pre-training performance in upstream, we analyze the model's perplexity (PPL) during the training process to measure its language modeling capability. 
 
We present the results in \cref{results:ppl}, where models marked as ``ours'' are fine-tuned on UnifiedCrawl-Amharic dataset using QLoRA. Both our fine-tuned models using QLoRA, XGLM-$564$M and XGLM-$4.5$B exhibit significantly lower perplexity to that of the original XGLM models. 

\begin{table}[H]
\begin{center}

\begin{tabular}{l c}
\toprule
\textbf{Models} & \textbf{PPL} \\ \midrule
XGLM-$564\mathrm{M}$ & 14,974.70 \\ 
XGLM-$564\mathrm{M}$ (ours) & \textbf{105.5} \\
\midrule
XGLM-$4.5\mathrm{B}$ & 35.6 \\
XGLM-$4.5\mathrm{B}$ (ours) & \textbf{19.6} \\ \bottomrule

\end{tabular}
\caption{Language Modeling Evaluation on Amharic}
\label{results:ppl}

 \end{center}
\end{table}

The original XGLM-$564\mathrm{M}$ model had a PPL of $14,974.7$, as it was not trained on Amharic. The perplexity was dramatically lowered to $105.6$ when trained on our UnifiedCrawl-Amharic dataset. Similarly, the PPL for the XGLM-$4.5\mathrm{B}$ model decreased from $35.6$ to $19.6$, indicating a $45\%$ improvement.

These results demonstrate fine-tuning the models using QLoRA on our dataset can lead to significant reductions in perplexity for across model sizes.

\subsubsection{Downstream Few Shot Prompting}
\label{res:downstream}

In downstream tasks, we compare the original model with the fine-tuned QLoRA model on the Amharic dataset under few-shot prompting. We report the F1 score and the EM (Exact Match) score for these evaluations. 

Few-shot performance comparisons between the original and fine-tuned models are shown in \cref{tab:downstream}, where models marked as ``ours'' are fine-tuned on UnifiedCrawl-Amharic dataset using QLoRA. Since the XGLM-$564\mathrm{M}$ was not pre-trained on Amharic, both F1 scores and EM scores are 0, and the score remained unchanged even after fine-tuning this model on our UnifiedCrawl-Amharic. The model is too small, and trained for too few tokens to perform reasonably for few-shot prompting.  

However, for the XGLM-$4.5\mathrm{B}$ model, the F1 score increased by $24\%$ from $8.0$ to $9.9$ after fine-tuning, and the EM score increased from $1.3$ to $2.3$.

This demonstrates fine-tuning specifically benefited the larger model, boosting its few-shot prompting performance on Question-Answering.

\begin{table}[H]
\begin{center}
\begin{tabular}{lcc}
\toprule
\textbf{Models} & \textbf{F1} & \textbf{EM} \\ \midrule
XGLM-$564\mathrm{M}$ & 0 & 0 \\
XGLM-$564\mathrm{M}$ (ours) & 0 & 0 \\
\midrule
XGLM-$4.5\mathrm{B}$ & 8.0 & 1.3 \\
XGLM-$4.5\mathrm{B}$ (ours) & \textbf{9.9} & \textbf{2.3} \\
\bottomrule
\end{tabular}
\caption{Few-shot Prompting Score on AmQA.}
\label{tab:downstream}

 \end{center}
\end{table}

\section{Ablation Studies}
\label{Chapter 6}
\label{sec:ablation}

We conduct ablation studies to analyze the impact of different modeling choices and validate the effectiveness of our approach. Specifically, we compare using full fine-tuning versus only adapting with lightweight QLoRA modules, examine trade offs between leveraging pre-trained versus randomly initialized models, and evaluate whether gains from pre-training on our UnifiedCrawl corpus translate to improved performance on downstream tasks.

\subsection{Comparison with Full Finetuning}
\label{sec:ablation:full_finetuning}
We compared the LM perplexity of our model, where we only trained a lightweight adapter QLoRA while keeping the original parameters frozen, to a fully fine-tuned model, where all the parameters of the model are trained without any adapters. We used our UnifiedCrawl-Amharic Dataset to train these models, in both cases. 

The results of this comparison is shown in \cref{tab:full_tuning}, where LM PPL is reported on Unified-Crawl-Amharic, and Few-shot F1/EM are on AmQA. Due to memory constraints on our GPU, we only performed full fine-tuning on the XGLM-$564\mathrm{M}$ model, as attempts to fully train a $4.5$B parameter model resulted in out-of-memory (OOM) errors.

We observe that fully fine-tuning the $564$M parameter model yields slightly better language modeling perplexity ($76.7$ vs. $105.6$) compared to using QLoRA to train adapters. However, full fine-tuning requires significantly more GPU memory and computational resources, i.e., it costs higher VRAM and compute compared to using QLoRA, for a relatively minor improvement. 

Furthermore, full-finetuning dramatically under-performs compared to using QLoRA on a larger model for the same compute - for example, the $564$M model achieves $76.7$ PPL with full-fine-tuning, whereas the $4.5$B model achieves $19.6$

We also evaluated the performance in a downstream few-shot prompting setting on the smaller model. We observed that the Few-shot prompting scores remained zero for both cases (i.e., Full Fine-tuning vs QLoRA) for the $564$M model, whereas the $4.5$B QLoRA model achieves F1 score of $9.9$. This further highlights the importance of using larger models with QLoRA.

\begin{table*}[ht]
\begin{center}
\begin{tabular}{llcc}
\toprule
\textbf{Model} & \textbf{LM PPL} & \textbf{Few-shot F1}  & \textbf{Few-shot EM} \\
\midrule
XGLM-$564\mathrm{M}$ (full finetune) & \textbf{76.7} & 0 & 0 \\  
XGLM-$564\mathrm{M}$ (ours) & 105.6 & 0 & 0 \\  
XGLM-$4.5\mathrm{B}$ (full finetune) & OOM & - & - \\ 
XGLM-$4.5\mathrm{B}$ (ours) & \textbf{19.6} & \textbf{9.9} & \textbf{2.3} \\ 
\bottomrule
\end{tabular}
\caption{Comparison of QLoRA with Full-fine-tuning}
\label{tab:full_tuning}
\end{center}
\end{table*}

\subsection{Comparison with Training from Scratch}

We also compared using QLoRA to adapt pre-trained models, to training a new model from scratch. For fair comparison, we use the same compute budget for all models as required for training the XGLM-$4.5\mathrm{B}$ model for $1$ epoch on UnifiedCrawl-Amharic. We train a ``base'' sized model with $110\mathrm{M}$ params, as well as a $74\mathrm{M}$ model (``compute-optimal'' model size, based on Chincilla~\cite{chinchilla} for our compute constraints).

The results are shown in \cref{tab:scratch}, where LM PPL is reported on Unified-Crawl-Amharic, and Few-shot F1/EM are on AmQA. We observed that at equal compute, our model(trained by using QLoRA) shows a substantial performance improvement compared to models that are trained from scratch (without QLoRA). From this, we conclude that training an already pre-trained models by using adapters are better than training a model from scratch, as these models can effectively utilize their prior knowledge gained from multi-lingual pre-training. 

\begin{table*}[ht]
\begin{center}
\begin{tabular}{lccc}
\toprule
\textbf{Model} & \textbf{LM PPL} & \textbf{Few-shot F1} & \textbf{Few-shot EM} \\
\midrule
GPT2-74M (scratch) & 105.2 & 1.2 & 0  \\ 
GPT2-110M (scratch) & 106.1 & 1.3 & 0  \\ 
\midrule
XGLM-$4.5\mathrm{B}$ (Ours) & \textbf{19.6} & \textbf{9.9} & \textbf{2.3} \\
\bottomrule
\end{tabular}
\caption{Comparison of QLoRA with training from scratch}
\label{tab:scratch}
\end{center}
\end{table*}

\subsection{Comparison on Downstream Supervised Training}
We also compare our models (fine-tuned on UnifiedCrawl-Amharic) with baseline models on downstream supervised learning on QA task (AmQA). We use QLoRA for all the models, as training $4.5$B model results in OOM. 

We present these results in \cref{tab:supervised}, where PPL, F1 and EM are on the AmQA dataset. Models marked here with ``QLoRA'' are the original pre-trained models, which we finetune on downstream task using QLoRA. Models marked as ``ours'' add an additional step of fine-tuning on UnifiedCrawl-Amharic before the downstream training. 

While the $564\mathrm{M}$ model shows improvements in all scores, the perplexity of both the $4.5$B models, the baseline model and the model trained on UnifiedCrawl-Amharic, are very comparable, and so is their F1 and EM score. 

The gains observed in Language Modeling and in Few-shot Prompting did not translate to gains on downstream supervised training.  While the PPL of the XGLM-$564\mathrm{M}$ improved from $99.4$ to $59.5$, the PPL for the XGLM-4.5 model fine-tuned on the UnifiedCrawl-Amharic remained the same to that of the original model. This could perhaps be due to limited size or quality of this downstream dataset, which is only $1600$ training samples from very few wikipedia articles.

\begin{table}[H]
\begin{center}
\begin{tabular}{lccc}

\toprule
\textbf{Models} & \textbf{PPL} &\textbf{F1} & \textbf{EM} \\ 
\midrule
XGLM-$564\mathrm{M}$(QLoRA) & 99.4 & 0.6 & 0.2\\
XGLM-$564\mathrm{M}$ (ours) & \textbf{59.2 }& \textbf{2.9} & \textbf{0.7}\\
\midrule
XGLM-$4.5\mathrm{B}$ (QLoRA) & 2.2 & 35.0 & 20.5 \\
XGLM-$4.5\mathrm{B}$ (ours) & 2.2 & 34.7 & 20 \\ 
\bottomrule

\end{tabular}
\caption{Supervised-Training Score on AmQA.}
\label{tab:supervised}

 \end{center}
\end{table}
\section{Limitations and Future Works}
\label{Chapter 7}
\label{sec:limiations}

While our data extraction approach proves effective for low-resource languages, its applicability to high-resource languages is constrained by prolonged extraction time and storage challenges due to their abundance. Visualization reveals that conventional evaluation metrics, such as F1 and EM, may not adequately capture the nuances in the relationship between ground truth and predicted answers, given the linguistic diversity across languages.

As a future research direction, the data collection pipeline could be expanded on additional low-resource languages beyond the ones mentioned in this work. Additionally, our approach can be enhanced to improve the quality and diversity of the extracted data. 
We also believe exploring alternative model architectures, such as BLOOM and mT5, during the fine-tuning stage holds promise for achieving enhanced practical deployment. Moreover, a more comprehensive evaluation across diverse downstream tasks is essential to validate real-world performance gains resulting from our extracted data, UnifiedCrawl, and the recommended model adaptation technique.

By addressing these research directions, we aim to develop a robust technique that effectively broadens the accessibility and capabilities of Large Language Models (LLMs) for low-resource languages. This approach contributes to the global democratization of Natural Language Processing (NLP) by making advanced language models more widely available and applicable.

\section{Conclusion}
\label{Chapter 8}
\label{sec:conclusion}

To summarize, our key contributions are two-fold: First, we introduced an efficient technique to aggregate and extract large monolingual datasets for low-resource languages from the entire Common Crawl corpus. By selectively filtering archived data and minimizing storage needs, we obtained raw text data larger than any existing sources using only consumer hardware. Second, we demonstrated effective adaptation of multilingual LLMs by fine-tuning lightweight adapter modules on our extracted datasets. Fine-tuning 4.5B parameter models with adapters using QLoRA resulted in significant perplexity reductions and gains in few-shot prompting scores on Amharic language, with less than 1 GPU-day of compute. Our method and source code make progress towards democratizing LLMs.

\bibliography{zs}
\bibliographystyle{acl_natbib}

\appendix

\section{Distribution of Languages in Common Crawl}
\label{appendix:A}

\subsection{Distribution of Languages in Common Crawl except English}
\label{appendix:lang_dist1}

\begin{figure}[H]
\begin{center}
        \begin{tikzpicture}[scale=0.9]
        \begin{axis}[%
            xtick={0, 5, 10, 15, 20, 25, 30, 35, 40, 45, 50, 55, 60, 65, 70, 75, 80, 85, 90, 95, 100, 105, 110, 115, 120, 125, 130, 135, 140, 145, 150},
            xticklabels={eng, spa, por, swe, ell, fin, srp, est, msa, isl, eus, kaz, nno, cym, tat, tgk, som, pus, fry, cos, zul, xho, mri, bih, sna, ina, ile, lug, crs, run, zha},
            xticklabel style = {font=\tiny,rotate=90},
            yticklabel style = {font=\tiny},
            label style={font=\tiny},
            xlabel={Languages ->},
            ylabel={Percentage of Data (Log Scale) ->},
            ytick distance=0.5,
            ymin=0,
            axis x line=bottom,
            axis y line=left,
            ymajorgrids=true,
            grid style=dashed,
            xmajorticks=true,
            scaled y ticks=false,
            yticklabel=\pgfkeys{/pgf/number format/.cd,fixed,precision=1,zerofill}\pgfmathprintnumber{\tick}\%,
            ]
            \addplot[red] table[x expr={\thisrow{x}>0?\thisrow{x}:nan},y=percent] {data/language_cc.dat};
        \end{axis}
    \end{tikzpicture}
    \caption{Distribution of Languages in Common Crawl except English}
    \label{fig:lang_dist1}
    \end{center}
\end{figure}
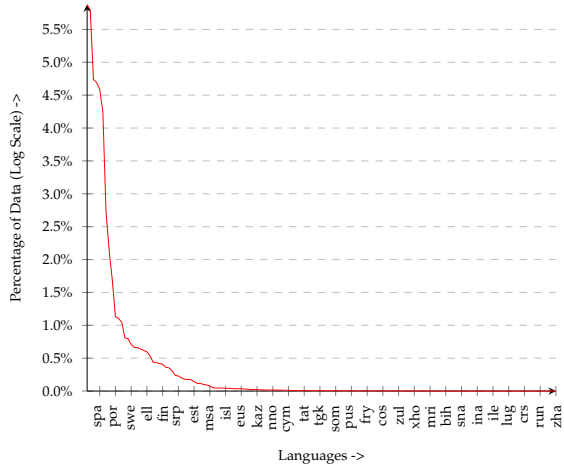

\subsection{ Distribution of Languages in Common Crawl after Top 60}
\label{appendix:lang_dist2}

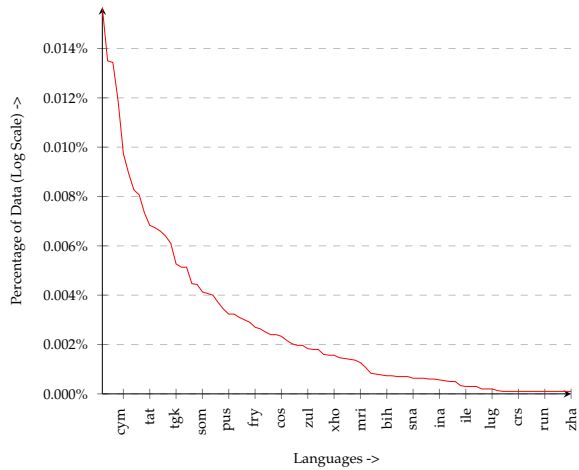
\begin{figure}[H]
\begin{center}
        \begin{tikzpicture}[scale=0.9]
        \begin{axis}[%
            xtick={0, 5, 10, 15, 20, 25, 30, 35, 40, 45, 50, 55, 60, 65, 70, 75, 80, 85, 90, 95, 100, 105, 110, 115, 120, 125, 130, 135, 140, 145, 150},
            xticklabels={eng, spa, por, swe, ell, fin, srp, est, msa, isl, eus, kaz, nno, cym, tat, tgk, som, pus, fry, cos, zul, xho, mri, bih, sna, ina, ile, lug, crs, run, zha},
            xticklabel style = {font=\tiny,rotate=90},
            yticklabel style = {font=\tiny},
            label style={font=\tiny},
            xlabel={Languages ->},
            ylabel={Percentage of Data (Log Scale) ->},
            ytick distance=0.002,
            ymin=0,
            axis x line=bottom,
            axis y line=left,
            ymajorgrids=true,
            grid style=dashed,
            xmajorticks=true,
            scaled y ticks=false,
            yticklabel=\pgfkeys{/pgf/number format/.cd,fixed,precision=3,zerofill}\pgfmathprintnumber{\tick}\%,
            ]
            \addplot[red] table[x expr={\thisrow{x}>60&&\thisrow{x}<154?\thisrow{x}:nan},y=percent] {data/language_cc.dat};
        \end{axis}
    \end{tikzpicture}
    \caption{Distribution of Languages in Common Crawl after Top 60}
    \label{fig:lang_dist2}
    \end{center}
\end{figure}

\begin{table*}[htb]
\centering
\begin{tabular}{l l l l}
\toprule
\textbf{Model Type} & \textbf{Multilingual LLMs} & \textbf{Size (\# Params)} & \textbf{\# Languages} \\ 
\midrule
\multirow{3}{*}{\textbf{Encoder-Only}} & mBERT~\cite{bert} & 180M & 104 \\ 
 & XLM-R~\cite{xlmr} & 225M-10.7B & 15/100 \\ 
 & XY-LENT~\cite{xylent} & 480M-2.1B & 21 \\
\midrule
\multirow{6}{*}{\textbf{Decoder-Only}} & XGLM~\cite{xglm} & 540M-7.5B & 30/134 \\
 & mGPT~\cite{mgpt} & 1.3B & 101 \\ 
 & PaLM~\cite{palm} & 540B & 122 \\
 & BLOOM~\cite{bloom} & 560M-175B & 46 \\
 & BLOOMZ~\cite{bloomz} & 560M-175B & 46 \\
 & GPT-3~\cite{gpt3} & 175B & 1 \\ 
\midrule
\multirow{3}{*}{\textbf{Encoder-Decoder }} & mT5~\cite{mT5} & 580M-13B & 101 \\
 & mT0~\cite{bloomz} & 580M-13B & 101 \\
 & mBART~\cite{mbart} & 680M & 25 \\ \bottomrule
\end{tabular}
\caption{Overview of Multilingual LLMs}
\label{tab:llm_size}
\end{table*}

\section{Overview of Multilingual LLMs}
\label{appendix:llm_size}
See \cref{tab:llm_size}

\end{document}